\documentclass[10pt,twocolumn,letterpaper]{article}

\usepackage{cvpr}
\usepackage{times}
\usepackage{epsfig}
\usepackage{graphicx}
\usepackage{amsmath}
\usepackage{amssymb}
\usepackage{booktabs}
\usepackage{caption} 
\usepackage[caption=false,font=footnotesize]{subfig}

\usepackage[utf8]{inputenc}
\usepackage{color}

\usepackage[pagebackref=true,breaklinks=true,letterpaper=true,colorlinks,bookmarks=false]{hyperref}

\cvprfinalcopy % *** Uncomment this line for the final submission

\def\cvprPaperID{1721} % *** Enter the CVPR Paper ID here

\ifcvprfinal\fi
\newcommand*\samethanks[1][\value{footnote}]{\footnotemark[#1]}
\begin{document}
\newcommand*{\affaddr}[1]{#1}
\newcommand*{\affmark}[1][*]{\textsuperscript{#1}}

\title{Self-supervised learning of visual features through embedding images into text topic spaces}

% \author{Lluis Gomez\affmark[1]\thanks{Equal Contribution} \and Yash Patel\affmark[2]\samethanks \and Marçal Rusiñol\affmark[1] \and C.V Jawahar\affmark[2] \and Dimosthenis Karatzas\affmark[1]\\
% \affaddr{\affmark[1]Computer Vision Center, Universitat Autònoma de Barcelona, Spain}\\ 
% \affaddr{\affmark[2]CVIT, KCIS, IIIT Hyderabad, India}\\
% {\tt\small \{lgomez, marcal, dimos\}@cvc.uab.es, yash.patel@students.iiit.ac.in, jawahar@iiit.ac.in}
% }
\author{Lluis Gomez\thanks{These authors contributed equally to this work}\\
Computer Vision Center, UAB, Spain\\
{\tt\small lgomez@cvc.uab.es}
\and
Yash Patel\samethanks\\
CVIT, KCIS, IIIT Hyderabad, India\\
{\tt\small yash.patel@students.iiit.ac.in}
\and
Marçal Rusiñol\\
Computer Vision Center, UAB, Spain\\
{\tt\small marcal@cvc.uab.es}
\and
Dimosthenis Karatzas\\
Computer Vision Center, UAB, Spain\\
{\tt\small dimos@cvc.uab.es}
\and
C.V. Jawahar\\
CVIT, KCIS, IIIT Hyderabad, India\\
{\tt\small jawahar@iiit.ac.in}
}

\maketitle
%\thispagestyle{empty}

%%%%%%%%% ABSTRACT
\begin{abstract}
End-to-end training from scratch of current deep architectures for new computer vision problems would require Imagenet-scale datasets, and this is not always possible.

%{\color{red}Supervised deep learning algorithms are insanely data-hungry. %Providing enough data to reach state-of-the art performance may require a %prohibitive amount of human annotation effort.}

In this paper we present a method that is able to take advantage of freely available multi-modal content to train computer vision algorithms without human supervision. We put forward the idea of performing self-supervised learning of visual features by mining a large scale corpus of multi-modal (text and image) documents. We show that discriminative visual features can be learnt efficiently by training a CNN to predict the semantic context in which a particular image is more probable to appear as an illustration. For this we leverage the hidden semantic structures discovered in the text corpus with a well-known topic modeling technique.
%For this we leverage the hidden semantic structures discovered by a well-known topic modeling technique over the text corpus.

Our experiments demonstrate state of the art performance in image classification, object detection, and multi-modal retrieval compared to recent self-supervised or natural-supervised approaches.
%In addition, we show how image retrieval can be seamlessly performed in either the visual and the textual domain, as a direct consequence of embedding images into the text-learnt topic space.
%Moreover we can do image retrieval from textual queries , this is something novel in self-supervised learning
\end{abstract}

%%%%%%%%% BODY TEXT

%-------------------------------------------------------------------------
\section{Introduction}

A picture is worth a thousand words. When we read an article about an unknown object, event, or place we greatly appreciate that it is accompanied by some image that supports the textual information. These images complement the textual description and at the same time provide context to our imagination. Illustrated texts are thus ubiquitous in our culture: newspaper articles, encyclopedia entries, web pages, etc. Can we take advantage of all this available multi-modal content to train computer vision algorithms without human supervision?

\begin{figure}
  \includegraphics[width=\columnwidth]{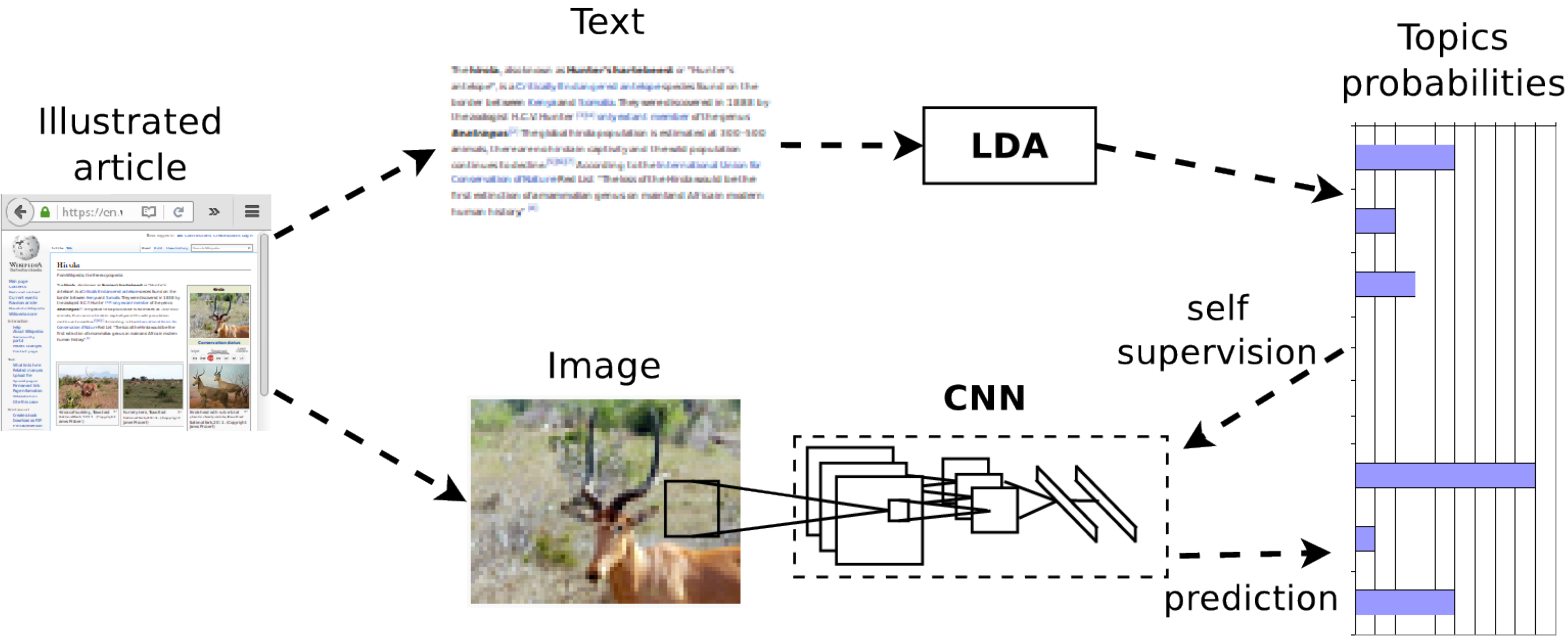}
    \caption{Our CNN learns to predict the semantic context in which images appear as illustration. Given an illustrated article we project its textual information into the topic-probability space provided by a topic modeling framework. Then we use this semantic level representation as the supervisory signal for CNN training.}
  \label{fig:method}
\end{figure}

\begin{figure}
\centering
  \subfloat[]{\label{subfig:a} \includegraphics[width=0.23\textwidth]{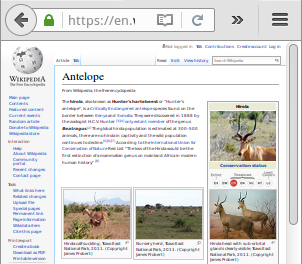}}
  \hfill
  \subfloat[]{\label{subfig:b} \includegraphics[width=0.23\textwidth]{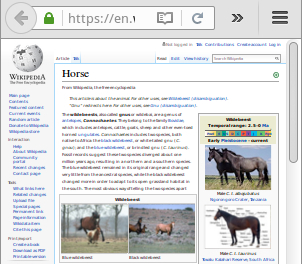}}\\
  \subfloat[]{\label{subfig:c} \includegraphics[width=0.475\textwidth]{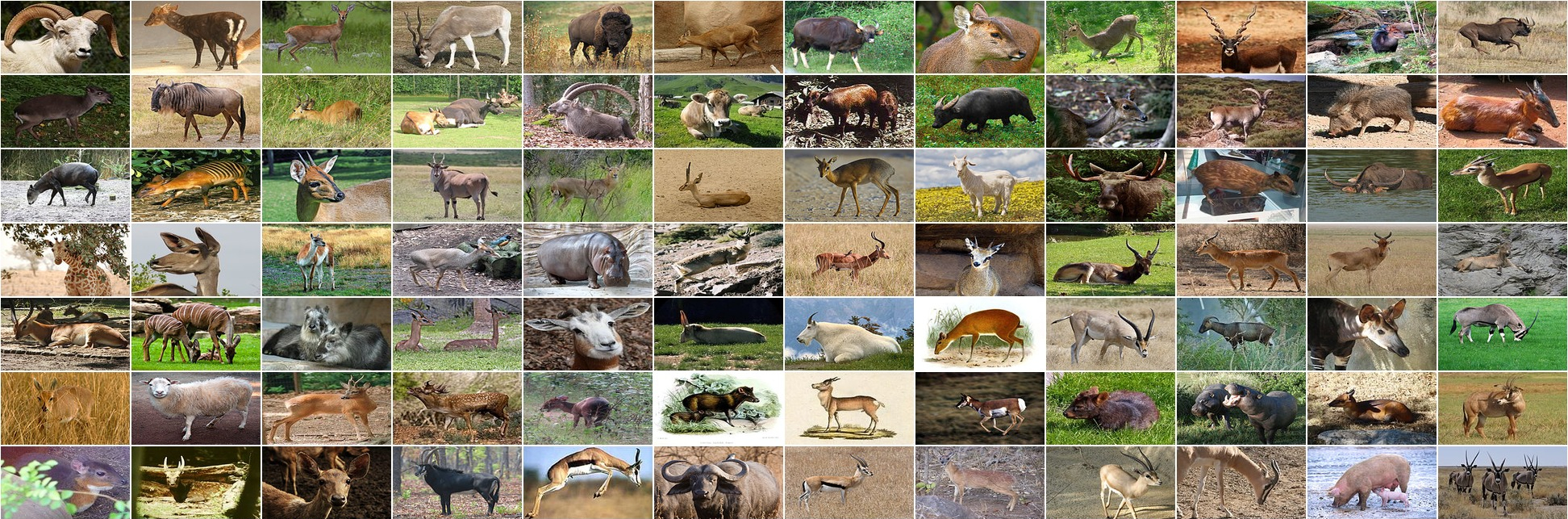}}
  %\caption{Training data used for self-supervised learning : Illustrated Wikipedia articles about specific entities, like ``Antelope''(a) or ``Horse''(b). Each article typically contains around half a dozen images. The total number of images for broader topics, e.g. ``herbivorous mammals''(c), can easily reach hundreds or thousands.}
    \caption{Illustrated Wikipedia articles about specific entities, like ``Antelope'' (a) or ``Horse'' (b), typically contain around five images. The total number of images for broader topics, e.g. ``herbivorous mammals'' (c), can easily reach hundreds or thousands.}
  \label{visual_topics}
\end{figure}

Training deep networks requires a signficant amount of annotated data. The emergence of large-scale annotated datasets~\cite{deng2009imagenet} has undoubtedly been one of the key ingredients for the tremendous impact deep learning is having on almost every computer vision task. However, the amount of human resources needed to manually annotate such datasets represents a problem. The goal of this paper is to propose an alternative solution to fully supervised training of CNNs by leveraging the correlation between images and text found in illustrated articles.

In most cases human generated data annotations consist of textual information with different granularity depending on the visual task they address: a single word to identify an object/place (classification), a list of words that describe the image (labeling), or a %explanatory 
descriptive phrase of the scene shown (captioning). In this paper we consider that text found in illustrated articles can be leveraged as a type of image annotation, albeit being a very noisy one. The key benefit of this approach is that these annotations can be obtained for ``free''.

Recent work in self-supervised or natural-supervised learning for computer vision has demonstrated success in using non-visual information as a form of self-supervision for visual feature learning ~\cite{agrawal2015learning,wang2015unsupervised,doersch2015unsupervised,owens2016ambient}. Surprisingly, the textual modality has been ignored until now in self-supervised methods for CNN training.

In this paper we present a method that performs self-supervised learning of visual features by mining a large scale corpus of multi-modal web documents (Wikipedia articles). We claim that it is feasible to learn discriminative features by training a CNN to predict the semantic context in which a particular image is more probable to appear as an illustration. For this we represent textual information at the topic level, by leveraging the hidden semantic structures discovered by the Latent Dirichlet Allocation (LDA) topic modeling framework~\cite{blei2003latent}, and use this representation as supervision for visual learning as shown in Figure~\ref{fig:method}. 

As illustrated in Figure~\ref{visual_topics}, the intuition behind using topic-level text descriptors is that the amount of visual data available about specific objects (e.g. a particular animal) is limited in our data collection, while it would be easy to find enough images representative of broader object categories (e.g. ``mammals''). As a result of this approach the expected visual features that we are going to learn will be generic for a given topic, but still useful for other, more specific, computer vision tasks.

{Our main motivation is to explore how strong are language semantics as a supervisory signal to learn visual features. In this paper we demonstrate that CNNs can learn rich features from noisy and unstructured textual annotations. By training a CNN to directly project images into a textual semantic space, our method is not only able to learn visual features from scratch without a large annotated dataset, but it can also perform multi-modal retrieval in a natural way without any extra annotation or learning efforts.}

The contributions of this paper are the following: First, we present a method that performs self-supervised feature learning of visual features by leveraging the correlation between images and the semantic context in which they appear. Second, we  experimentally demonstrate that the learned visual features provide comparable or better performance to recent self-supervised and unsupervised algorithms in image classification, object detection, and multi-modal retrieval tasks on standard benchmarks. %{\color{blue}Finally, we demonstrate that our method naturally learns to perform multi-modal retrieval (image-text) and clearly outperforms unsupervised approaches, and has competitive performance to supervised approaches on standard benchmark.}
%Our findings ... 

%Third, we show a novel application that opens the door to do generic image retrieval from both image and textual queries.

%-------------------------------------------------------------------------
\section{Related Work}
\label{related_work}
Work in unsupervised data-dependent methods for learning visual features has been mainly focused on algorithms that learn filters one layer at a time. A number of unsupervised algorithms have been proposed to that effect, such as sparse-coding, restricted Boltzmann machines (RBMs), auto-encoders~\cite{zhao2015stacked}, and K-means clustering~\cite{coates2010analysis,dundar2015convolutional,krahenbuhl2015data}. However, despite the success of such methods in several unsupervised learning benchmark datasets, a generic unsupervised method that works well with real-world images does not exist.

As an alternative to fully-unsupervised algorithms, there has recently been a growing interest in self-supervised or natural-supervised approaches that make use of non-visual signals, intrinsically correlated to the image, as a form to supervise visual feature learning. Agrawal \etal~\cite{agrawal2015learning} make use of egomotion information obtained by odometry sensors mounted on a vehicle to pre-train a CNN model.  Wang \& Gupta~\cite{wang2015unsupervised} use relative motion of objects in videos by leveraging the output of a tracking algorithm. Doersch \etal~\cite{doersch2015unsupervised} learn visual features by predicting the relative position of image patches within the image. In Owens \etal~\cite{owens2016ambient} the supervisory signal comes  from  a  modality  (sound)  that  is  complementary  to vision. 

In this paper we explore a different modality, text, for self-supervision of CNN feature learning. As mentioned earlier, text is the default choice for image annotation in many computer vision tasks. This includes classical image classification~\cite{deng2009imagenet,everingham2010pascal},  annotation~\cite{duygulu2002object,huiskes2008mir}, and captioning~\cite{Ordonez2011im2text,lin2014microsoft}. In this paper, we extend this to a larger level of abstraction by capturing text semantics with topic models. Moreover, we avoid using any human supervision by leveraging the correlation between images and text in a largely abundant corpus of illustrated web articles. %the semantic context in which they appear

% Other LDA based methods for Annotation and retrieval tasks.
Our method is closely related with various image retrieval and annotation algorithms that also use a topic modeling framework in order to embed text and images in a common space. Multi-modal LDA (mmLDA) and correspondence LDA (cLDA)~\cite{blei2003modeling} methods learn the joint distribution of image features and text captions by finding correlations between the two sets of hidden topics. Supervised variations of LDA are presented in ~\cite{rasiwasia2013latent,wang2011max,putthividhy2010topic} where the discovered topics are driven by the semantic regularities of interest for the classification task. Sivic \etal~\cite{sivic2005discovering} adopt BoW representation of images for discovering objects in images using pLSA~\cite{hofmann2001unsupervised} for topic modelling. Feng \etal~\cite{feng2010topic} uses the joint BoW representation of text and image for learning LDA. 
Most cross-modal retrieval methods work with the idea of representing data of different modalities into a common space where data related to same topic of interest tend to appear together. The unsupervised methods in this domain utilize co-occurrence information to learn a common representation across different modalities. Verma \etal~\cite{verma2014im2text} do image-to-text and text-to-image retrieval using LDA~\cite{blei2003latent} for data representation. Methods such as those presented in ~\cite{rasiwasia2010new, gong2014multi,pereira2014role,li2011face} use Canonical Correlation Analysis (CCA) for establishing relationships between data of different modalities. Rasiwasia \etal~\cite{rasiwasia2010new} proposed a method for cross-modal retrieval by representing text using LDA~\cite{blei2003latent}, image using BoW and CCA for finding correlation across different modalities.

Our method is related to these image annotation and image retrieval methods in the sense that we use LDA~\cite{blei2003latent} topic-probabilities as common representation for both image and text. However, we differ from all these methods in that we use the topic level representations of text to supervise the visual feature learning of a convolutional neural network. Our CNN model, by learning to predict the semantic context in which images appear as illustrations, learns generic visual features that can be leveraged for other visual tasks. A similar idea is explored in the work of Gordo and Larlus~\cite{gordo2017} in these same proceedings, where image captions are leveraged to learn a global visual representation for semantic retrieval.

%-------------------------------------------------------------------------
\section{TextTopicNet}
\label{sec:method}

In order to train a CNN to predict semantic context from images (TextTopicNet) we propose a two-fold method: First, we learn a topic model on a text corpus of a dataset composed by pairs of correlated texts and images (i.e. illustrated articles). Second, we train a deep CNN model to predict text representations (topic-probabilities) directly from the image pixels. Figure~\ref{fig:method} shows a diagram of the method.

\subsection{LDA topic modeling}
\label{sec:train_lda}

Our self-supervised learning framework assumes that the textual information associated with the images in our dataset is generated by a mixture of hidden topics. Similar to various image annotation and image retrieval methods discussed in \ref{related_work}, we make use of the Latent Dirichlet Allocation (LDA) algorithm~\cite{blei2003latent} for discovering those latent topics and representing the textual information associated with a given image as a probability distribution over the set of discovered topics. 

Representing text at topic level instead of at word level (BoW) provides us with: (1) a more compact representation (dimensionality reduction), and (2) a more semantically meaningful interpretation of descriptors. 

%- TODO Elaborate with (i) one or two more sentences (ii) one worked out example (discussing Figure 1 or an appropriate figure). 

LDA is a generative statistical model of a text corpus where each document can be viewed as a mixture of various topics, and each topic is characterized by a probability distribution over words. LDA can be represented as a three level hierarchical Bayesian model. Given a text corpus consisting of $M$ documents and a dictionary with $N$ words, Blei \etal define the generative process~\cite{blei2003latent} for a document $d$ as follows:

\begin{itemize}
\item{Choose $\theta \sim Dirichlet(\alpha)$.}
\item{For each of the $N$ words $w_n$ in $d$:}
 \begin{itemize}
 \item{Choose a topic $z_n \sim Multinomial(\theta)$.}
 \item{Choose a word $w_n$ from $P(w_n \mid z_n, \beta)$, a multinomial probability conditioned on the topic $z_n$.}
 \end{itemize}
\end{itemize}

\noindent
where $\theta$ is the mixing proportion and is drawn from a Dirichlet prior with parameter $\alpha$, and both $\alpha$ and $\beta$ are corpus level parameters, sampled once in the process of generating a corpus. Each document is generated according to the topic proportions $z_{1:K}$ and word probabilities over $\beta$. The probability of a document $d$ in a corpus is defined as : 

\small
\begin{equation}
P(d\mid\alpha, \beta) =  \nonumber
\int_{\theta}P(\theta \mid\alpha)\left(\prod_{n=1}^{N}\sum_{z_{K}}^{ } P(z_{K} \mid \theta)P(w_{n}\mid z_{K},\beta)\right)d\theta \nonumber
\end{equation}
\normalsize

Learning LDA~\cite{blei2003latent} on a document corpus provides two sets of parameters: word probabilities given topic $P(w\mid z_{1:K})$ and topic probabilities given document $P(z_{1:K} \mid d)$. Therefore each document is represented in terms of topic probabilities $z_{1:K}$ (being $K$ the number of topics) and word probabilities over topics. Any new (unseen) document can be represented in terms of a probability distribution over the topics of the learned LDA model by projecting it into the topic space.

% TODO : Add LDA equation. Define the notations. Discuss why we pick LDA as the preferred topic model.

\subsection{Training a CNN to predict semantic topics}
\label{sec:train_cnn}

\begin{figure*}
  \includegraphics[width=\textwidth]{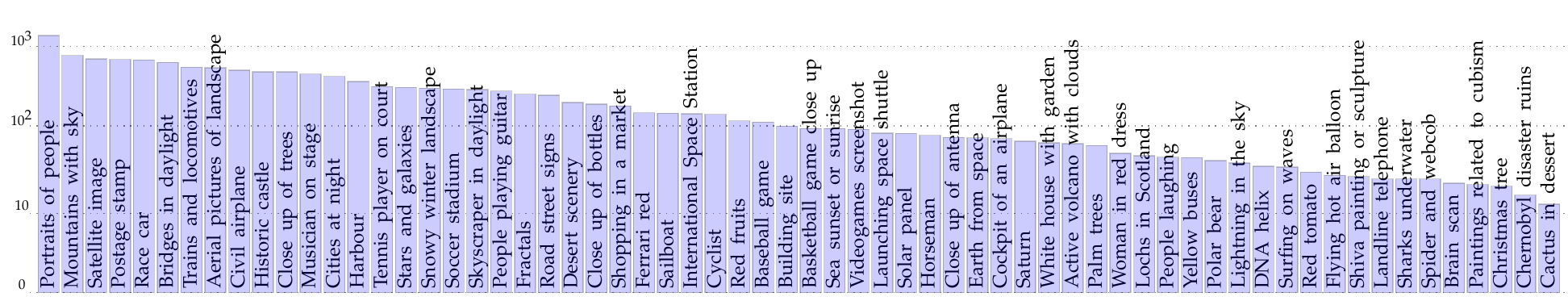}
    \caption{Number of relevant images (log scale) for a variety of semantic queries on the ImageCLEF Wikipedia collection~\cite{tsikrika2011overview}.}
  \label{fig:dataset_analysis}
\end{figure*}

\begin{figure*}
  \def\svgwidth{\textwidth}\input{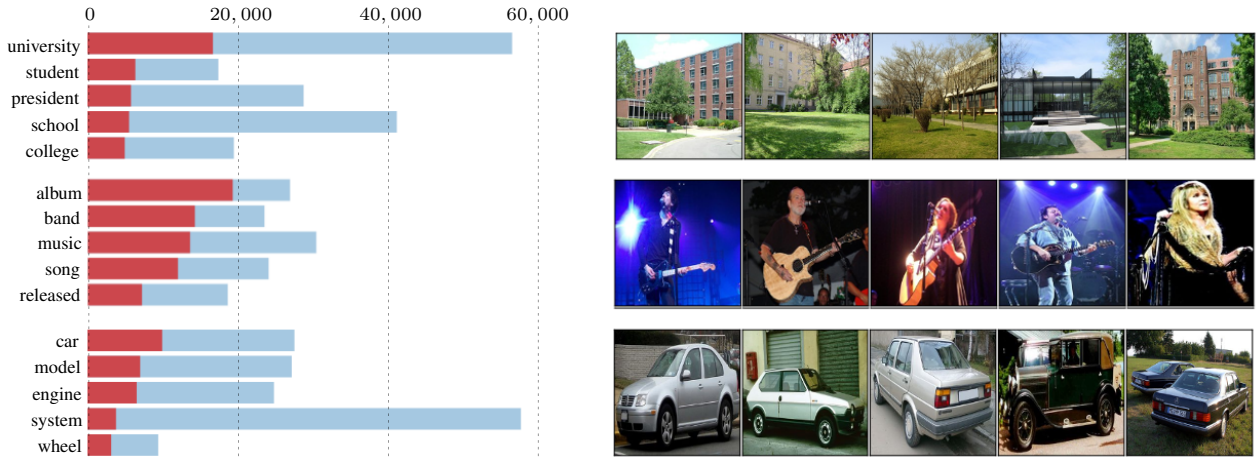}
    \caption{Top-5 most relevant words for 3 of the discovered topics by LDA analysis (left), and top-5 most relevant images for the same topics (right). Overall word frequency is shown in blue, and estimated word frequency within the topic in red.}
  \label{fig:topic_words_imgs}
\end{figure*}

We train a CNN to predict text representations (topic probability distributions) from images. Our intuition is that we can learn useful visual features by training the CNN to predict the semantic context in which a particular image is more probable to appear as an illustration.

For our experiments we make use of two different architectures. One is the 8 layers CNN CaffeNet~\cite{jia2014caffe}, a replication of the AlexNet~\cite{krizhevsky2012imagenet} model with some differences (it does not train with the relighting data-augmentation, and the order of pooling and normalization layers is switched). The other architecture is a 6 layers CNN resulting from removing the 2 first convolutional layers from CaffeNet. This smaller network is used to do experiments with tiny images.
%TODO add figure for topic net

For learning to predict the target topic probability distributions we minimize a sigmoid cross-entropy loss on our image dataset. We use a Stochastic Gradient Descent (SGD) optimizer, with base learning rate of  $0.001$, multiplied by $0.1$ every $50,000$ iterations, and momentum of $0.9$. The batch size is set to $64$. With these settings the network converges after $120,000$ iterations.
%mention grid search
% training time is about X hours on a Titan X GPU for CaffeNet.

We train our models on a subset of Wikipedia articles provided in the Wikipedia ImageCLEF dataset~\cite{tsikrika2011overview}. The  ImageCLEF  2010  Wikipedia  collection consists of $237,434$ Wikipedia images and the Wikipedia articles that contain these images.
{An important observation is that the data collection and filtering is not semantically driven. The original ImageCLEF dataset contains all Wikipedia articles which have versions in three languages (English, German and French) and are illustrated with at least one image in each version. Thus, we have a broad distribution of semantic subjects, similar as to the entire Wikipedia or other general-knowledge data collections. A semantic analysis of the data, extracted from the ground-truth of relevance assessments for the ImageCLEF retrieval queries, is shown in Figure~\ref{fig:dataset_analysis}.}
Although the dataset provides also human-generated annotations in this paper we train CNNs from scratch using only the raw Wikipedia articles and their images.

We consider only the English articles of the ImageCLEF Wikipedia collection. We also filter small images ($< 256$ pixels) and images with formats other than JPG (Wikipedia stores photographic images as JPG, and uses other formats for digital-born imagery). This way our training data is composed of $100,785$ images and $35,582$ unique articles. We use data augmentation by random crops and mirroring.
%- TODO Add Sec 3.3: Given the steps in Sec 3.1 and Sec 3.2, how we solve the original problem stated in Sec 1 (say self supervised learning.)

Figure~\ref{fig:topic_words_imgs} shows the top-5 most relevant words for three of the discovered topics by LDA analysis, and the top-5 most relevant images for such topics. We appreciate that the discovered topics correspond to broad semantic categories for which, a priori, it is difficult to find the most appropriate illustration. Still we observe that the most representative images for each topic present some regularities and thus allow the CNN to learn discriminative features, despite the noise introduced by other images that appear in articles from the same topic. 

On the other hand, a given image will rarely correspond to a single semantic topic. Because by definition the discovered topics by LDA have a certain semantic overlap. In this sense we can think of the problem of predicting topic probabilities as a multi-label classification problem in which all classes exhibit a large intra-class variability. These intuitions motivate our choice of a sigmoid cross-entropy loss for predicting targets interpreted as topic probabilities instead of a one hot vector for a single topic.

\subsection{Self-supervised learning of visual features}

Once the TextTopicNet model has been trained following the steps in Section~\ref{sec:train_lda} and Section~\ref{sec:train_cnn} it can be straightforwardly used in an image retrieval setting. Furthermore, it can be easily extended to an image annotation or captioning system by leveraging the common topic space in which text and images can be projected by the LDA and CNN models. 

However, in this paper we are more interested in analyzing the qualities of the visual features that we have learned by training the network to predict semantic topic distributions. We claim that the learned features, out of the common topic space, are not only of sufficient discriminative power but also carry more semantic information than features learned with other state of the art self-supervised and unsupervised approaches.

The proposed self-supervised learning framework will have thus a broad application in different computer vision tasks. With this spirit we propose the use of TextTopicNet as a convolutional feature extractor and as a CNN pre-training method. We evaluate these scenarios in the next section and compare the obtained results in different benchmarks with the state of the art.

%-------------------------------------------------------------------------
\section{Experiments}

% Class-wise mAP table goes here.
\begin{table*}[t]
\resizebox{\textwidth}{!}{
\begin{tabular}{l | c | c | c | c | c | c | c | c | c | c | c | c | c | c | c | c | c | c | c | c }
\toprule
Method &aer &bk &brd &bt &btl &bus &car &cat &chr &cow &din &dog &hrs &mbk &prs &pot &shp &sfa &trn &tv \\
\midrule
TextTopicNet (Wiki) &67 &44 &39 &53 &\textbf{20} &\textbf{49} &\textbf{68} &42 &43 &\textbf{33} &41 &\textbf{35} &70 &57 &82 &\textbf{30} &31 &\textbf{39} &65 &41 \\
Sound~\cite{owens2016ambient} &69 &\textbf{45} &38 &56 &16 &47 &65 &45 &41 &25 &37 &28 &\textbf{74} &\textbf{61} &\textbf{85} &26 &\textbf{39} &32 &\textbf{69} &38 \\
Texton-CNN &65 &35 &28 &46 &11 &31 &63 &30 &41 &17 &28 &23 &64 &51 &74 &9 &19 &33 &54 &30 \\
K-means &61 &31 &27 &49 &9 &27 &58 &34 &36 &12 &25 &21 &64 &38 &70 &18 &14 &25 &51 &25\\
Motion~\cite{wang2015unsupervised} &67 &35 &41 &54 &11 &35 &62 &35 &39 &21 &30 &26 &70 &53 &78 &22 &32 &37 &61 &34 \\
Patches~\cite{doersch2015unsupervised} &\textbf{70} &44 &\textbf{43} &\textbf{60} &12 &44 &66 &\textbf{52} &\textbf{44} &24 &\textbf{45} &31 &73 &48 &78 &14 &28 &\textbf{39} &62 &\textbf{43} \\
Egomotion~\cite{agrawal2015learning} &60 &24 &21 &35 &10 &19 &57 &24 &27 &11 &22 &18 &61 &40 &69 &13 &12 &24 &48 &28 \\
\midrule
ImageNet~\cite{krizhevsky2012imagenet} &79 &\textbf{71} &\textbf{73} &75 &\textbf{25} &60 &80 &\textbf{75} &51 &\textbf{45} &60 &\textbf{70} &\textbf{80} &\textbf{72} &\textbf{91} &42 &\textbf{62} &56 &82 &62 \\
Places~\cite{zhou2014learning} &\textbf{83} &60 &56 &\textbf{80} &23 &\textbf{66} &\textbf{84} &54 &\textbf{57} &40 &\textbf{74} &41 &\textbf{80} &68 &90 &\textbf{50} &45 &\textbf{61} &\textbf{88} &\textbf{63} \\
\bottomrule
\end{tabular}}
\caption{PASCAL VOC2007 per-class average precision (AP) scores for the classification task with pool5 features.}
\label{pascal_pool5_AP}
\end{table*}

In order to demonstrate the quality of the visual features learned by our text topic predictor (TextTopicNet) we have performed several experiments. First we analyze the quality of TextTopicNet top layers features for image classification on the PASCAL VOC2007 dataset~\cite{everingham2010pascal}. Second we compare our method with state of the art unsupervised learning algorithms for image classification on PASCAL and STL-10~\cite{coates2010analysis} datasets, and for object detection in PASCAL. Finally, we perform qualitative experiments on image retrieval from visual and textual queries.

% Experimental set-up for LDA goes here.
For all our experiments we make use of the same LDA topic model learned on a corpus of $35,582$ English Wikipedia articles from the ImageCLEF Wikipedia collection~\cite{tsikrika2011overview}. From the raw articles we remove stop-words and punctuation, and perform lemmatization of words. The word dictionary ($50,913$ words) is made from the processed text corpus by filtering those words that appear in less than $20$ articles or in more than $50\%$ of the articles. At the time of choosing the number of topics in our model we must consider that as the number of topics increase, the documents of the training corpus are partitioned into finer collections, and increasing the number of topics may also cause an increment on the model perplexity~\cite{blei2003latent}. Thus, the number of topics is an important parameter in our model. 
%The number of topics can affect the image classification results drastically, thus number of topics in LDA~\cite{blei2003latent} is an important parameter.
In the next section we take a practical approach and empirically determine the optimal number of topics in our model by leveraging validation data. 
%, in Figure~\ref{plot_map_num_topics}, we report performance of our method with varying number of topics for classification task on PASCAL VOC2007 dataset~\cite{everingham2010pascal}.

%The details of each dataset are provided in the supplementary material.

\subsection{Unsupervised feature learning for image classification}

In this experiment we evaluate how good are the learned visual features of the 6 layer CNN (CaffeNet) for image classification when trained with the self-supervised method explained in Section~\ref{sec:method}. Following~\cite{owens2016ambient} we extract features from top layers of the CNN and train one vs. rest linear SVMs for image classification in PASCAL VOC2007 dataset.

\begin{figure}[h]
  \includegraphics[width=\columnwidth]{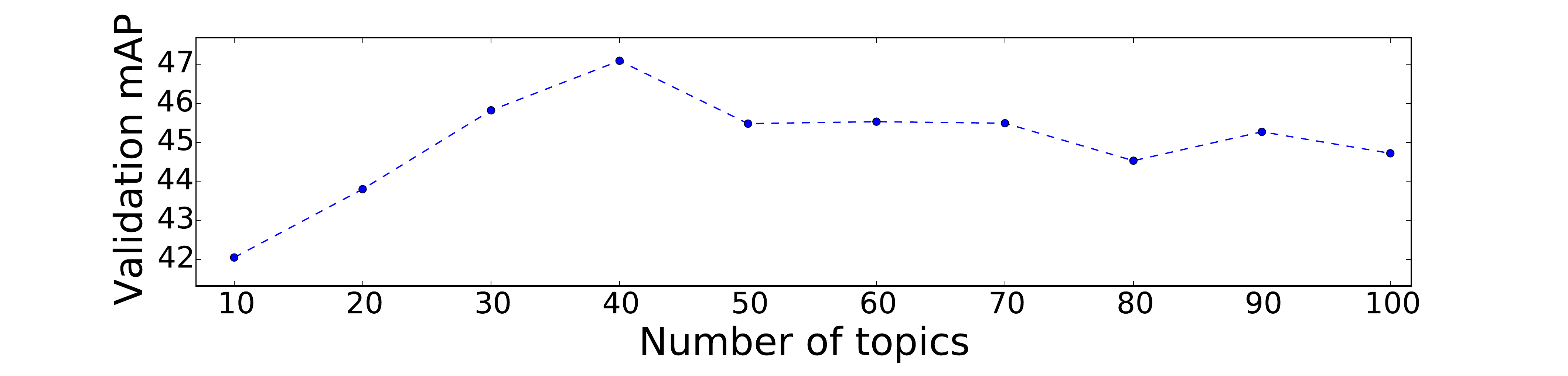}
    \caption{One vs. Rest linear SVM validation \%mAP on PASCAL VOC2007 by varying number of topics of LDA~\cite{blei2003latent} in our method.}
  \label{plot_map_num_topics}
\end{figure}

First of all, we perform model selection and parameter optimization using the standard train/validation split of the dataset. Figure~\ref{plot_map_num_topics} shows validation accuracy of SVM classification using \textit{fc7} features for different number of topics in our model. Best validation performance is obtained for $40$ topics. This configuration is kept for the rest of the experiments in this section.

Tables~\ref{pascal_pool5_AP} and~\ref{pascal_SVM_mAP} compare our results on the PASCAL VOC2007 test set with different state of the art self-supervised learning algorithms. Scores for all other methods are taken from~\cite{owens2016ambient}. We appreciate in Table~\ref{pascal_SVM_mAP} that using text semantics as supervision for visual feature learning outperforms all other modalities in this experiment. In Table~\ref{pascal_pool5_AP}, attention is drawn to the fact that our pool5 features are substantially more discriminative than the rest for the most difficult classes, see e.g. ``bottle'', ``pottedplant'' or ``cow''.

\begin{table}[h]
\begin{tabular}{ l | c | c | c | c}
\toprule
Method & max5 & pool5 & fc6 & fc7 \\
\midrule
TextTopicNet (Wiki) & - & \textbf{47.4} & \textbf{48.1} & \textbf{48.5} \\
Sound ~\cite{owens2016ambient} & \textbf{39.4} & 46.7 & 47.1 & 47.4 \\
Texton-CNN & 28.9 & 37.5 & 35.3 & 32.5 \\
K-means~\cite{krahenbuhl2015data} & 27.5 & 34.8 & 33.9 & 32.1 \\
Tracking~\cite{wang2015unsupervised} & 33.5 & 42.2 & 42.4 & 40.2 \\
Patch pos.~\cite{doersch2015unsupervised} & 26.8 & 46.1 & - & - \\
Egomotion~\cite{agrawal2015learning} & 22.7 & 31.1 & - & - \\
\midrule
TextTopicNet (COCO) & - & \textbf{50.7} & \textbf{53.1} & \textbf{55.4} \\
\midrule
ImageNet~\cite{krizhevsky2012imagenet} & \textbf{63.6} & \textbf{65.6} & \textbf{69.6} & \textbf{73.6} \\
Places~\cite{zhou2014learning} & 59.0 & 63.2 & 65.3 & 66.2 \\
\bottomrule
\end{tabular}
\caption{PASCAL VOC2007 \%mAP image classification.}
\label{pascal_SVM_mAP}
\end{table}

TextTopicNet (COCO) in Table~\ref{pascal_SVM_mAP} corresponds to a model trained with MS-COCO~\cite{lin2014microsoft} images and their ground-truth caption annotations as textual content. Since MS-COCO annotations are human generated, this entry can not be considered a self-supervised method, but rather as a kind of weakly supervised approach. Our interest in training this model is to show that having more specific textual content, like image captions, helps TextTopicNet to learn better features. In other words, there is an obvious correlation between the noise introduced in the self supervisory signal of our method and the quality of the learned features. Actually, the ImageNet entry in Table~\ref{pascal_SVM_mAP} can be somehow seen as a model with a complete absence of noise, i.e. each image corresponds exactly to one topic and each topic corresponds exactly to one class (a single word). Still, the TextTopicNet (Wiki) features, learned from a very noisy signal, perform surprisingly well compared with the ones of the TextTopicNet (COCO) model.

As an additional experiment we have calculated the 
classification performance of the combination of TextTopicNet (Wiki) and Sound entries in Table~\ref{pascal_SVM_mAP}. Here we seek insight about how complementary are the features learned with two different supervisory signals. By using the concatenation of \textit{fc7} features of those models the mAP increases to 54.81\%, indicating a certain degree of complementarity. 

We further analyze the qualities of the learned features by visualizing the receptive field segmentation of TextTopicNet convolutional units using the methodology of~\cite{zhou2014object,owens2016ambient}. The purpose of this experiment is to gain insight in what our CNN has learned to detect.

\begin{figure}
  \includegraphics[width=\columnwidth]{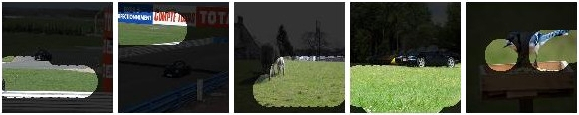}\\
  \includegraphics[width=\columnwidth]{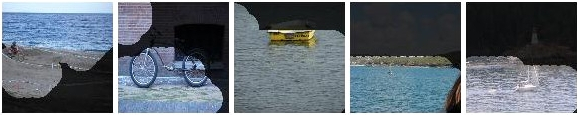}\\
  \includegraphics[width=\columnwidth]{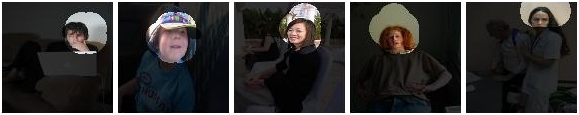}\\
  \includegraphics[width=\columnwidth]{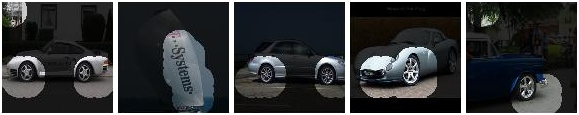}\\
  \includegraphics[width=\columnwidth]{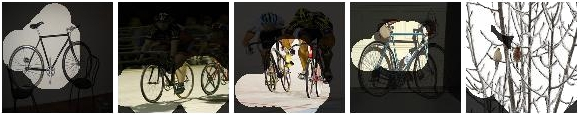}
  \caption{Top-5 activations for five units in \textit{fc7} layer of TextTopicNet (Wiki) model. While most TextTopicNet units are selective to generic textures, like grass or water, some of them are also selective for specific shapes, objects, and object-parts.}
  \label{rf_segmentations}
\end{figure}

Figure \ref{rf_segmentations} shows a selection of neurons in the \textit{fc7} layer of our model. We appreciate that our network units are quite generic, mainly selective to textures, shapes and object-parts, although some object-selective units are also present (e.g. faces). 

\subsection{Comparison to unsupervised pre-training and semi-supervised methods}

In this experiment we analyze the performance of TextTopicNet for image classification and object detection by fine-tuning the CNN weights to specific datasets (PASCAL and STL-10) and tasks.
	
For fine-tuning our network we use the following optimization strategy:  we use Stochastic Gradient
Descent (SGD) for $120,000$ iterations with an initial learning rate of $0.0001$ (reduced by $0.1$ every $30,000$ iterations), batch size of $64$, and momentum of $0.9$. We use data augmentation by random crops and mirroring. At test time we follow the standard procedure of averaging the net responses at $10$ random crops. For object detection we fine-tune our classification network using Fast R-CNN~\cite{girshick2015fast} with default parameters for $40,000$ iterations.

Table~\ref{pascal_finetuning_mAP} compares our results for image classification and object detection on PASCAL with different self-supervised learning algorithms.

\begin{table}[h]
\begin{tabular}{ l | c | c }
\toprule
Method &  classif. & detection \\
\midrule
TextTopicNet  & 55.7 & 43.0 \\
Sound ~\cite{owens2016ambient} & - & 44.1 \\
K-means~\cite{krahenbuhl2015data} & 56.6 & 45.6 \\
Tracking~\cite{wang2015unsupervised} & \textbf{62.8} & \textbf{47.4} \\
Patch pos.~\cite{doersch2015unsupervised} & 55.3 & 46.6 \\
Egomotion~\cite{agrawal2015learning} & 52.9 & 41.8 \\
%\midrule
%TextTopicNet (COCO) & - & - \\
\midrule
ImageNet~\cite{krizhevsky2012imagenet} & \textbf{69.6} & \textbf{73.6} \\
\midrule
%TextTopicNet + K-means~\cite{krahenbuhl2015data} & - & - \\
Egomotion~\cite{agrawal2015learning} + K-means~\cite{krahenbuhl2015data}& 54.2 & 43.9 \\
Tracking~\cite{wang2015unsupervised} + K-means~\cite{krahenbuhl2015data}& 63.1 & 47.2 \\
Patch pos.~\cite{doersch2015unsupervised} + K-means~\cite{krahenbuhl2015data}& 65.3 & 51.1 \\
\bottomrule
\end{tabular}
\caption{PASCAL VOC2007 finetuning \%mAP for image classification and object detection.}
\label{pascal_finetuning_mAP}
\end{table}

% TODO : add some introduction to pass from one dataset to another?

Table~\ref{stl10_finetuning_Acc} compares our classification accuracy on STL-10 with different state of the art unsupervised learning algorithms. In this experiment we make use of the shortened 6 layers network in order to adapt better to image sizes for this dataset ($96\times96$ pixels). We do fine-tuning with the same hyper-parameters as for the 6 layer network.

The standard procedure on STL-10 is to perform unsupervised training on a provided set of $100,000$ unlabeled images, and then supervised training on the labeled data. While our method does not directly compare with unsupervised and semi-supervised methods in Table~\ref{stl10_finetuning_Acc}, because of the distinct approach (self-supervision), the experiment  provides insight about the added value of self-supervision compared with fully-unsupervised data-driven algorithms. It is important to notice that we do not make use of the STL-10 unlabeled data in our training.

\begin{table}[h]
\centering
\begin{tabular}{ l | c  }
\toprule
Method &  Acc. \\
\midrule
TextTopicNet (Wiki) - CNN-finetuning *& \textbf{76.51\%} \\
TextTopicNet (Wiki) - fc7+SVM *& 66.00\% \\
\midrule
Semi-supervised auto-encoder~\cite{zhao2015stacked} & \textbf{74.33}\% \\
Convolutional k-means~\cite{dundar2015convolutional} & 74.10\% \\
CNN with Target Coding~\cite{yang2015deep} & 73.15\% \\
Exemplar convnets~\cite{dosovitskiy2014discriminative} & 72.80\%\\
Unsupervised pre-training~\cite{paine2014analysis} & 70.20\% \\
Swersky \etal~\cite{swersky2013multi} * & 70.10\%\\
C-SVDDNet~\cite{wang2016unsupervised} & 68.23\% \\ 
K-means (Single layer net)~\cite{coates2010analysis} & 51.50\% \\
Raw pixels & 31.80\% \\
\bottomrule
\end{tabular}
\caption{STL-10 classification accuracy. Methods with an asterisk mark make use of external (unlabeled) data.}
\label{stl10_finetuning_Acc}
\end{table}

\begin{figure*}
\centering
  \includegraphics[width=0.3\textwidth]{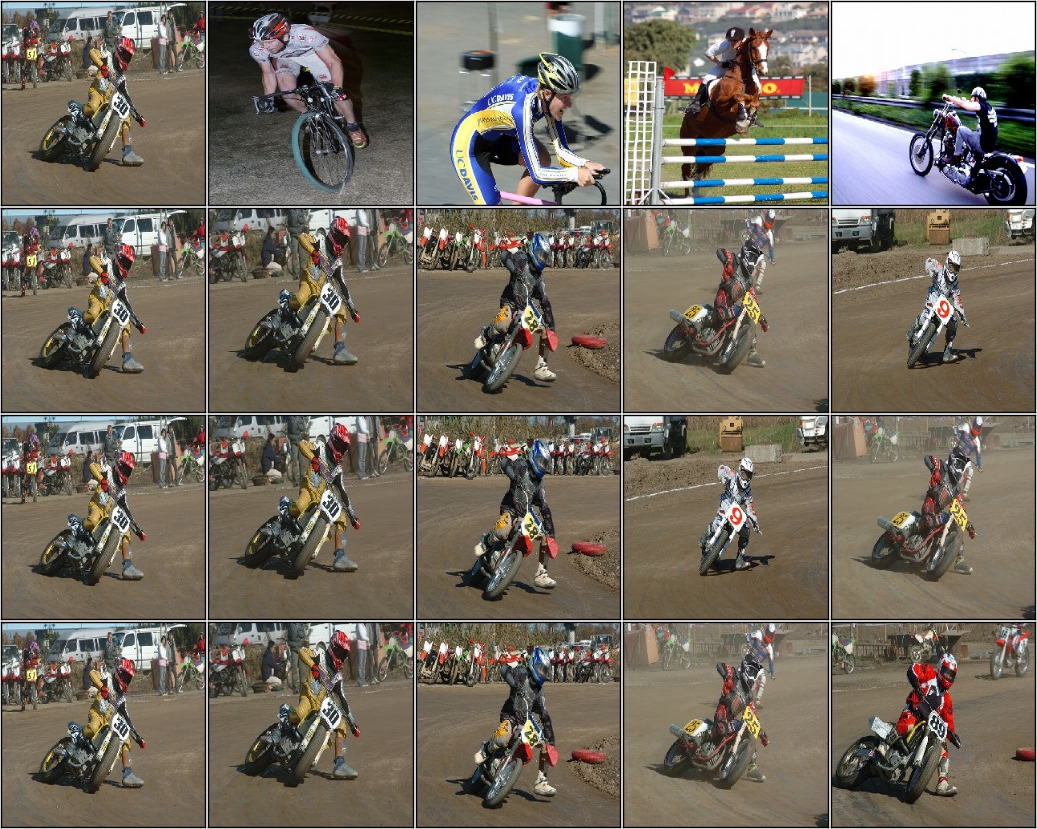}
  \hspace{1pt}
  \includegraphics[width=0.3\textwidth]{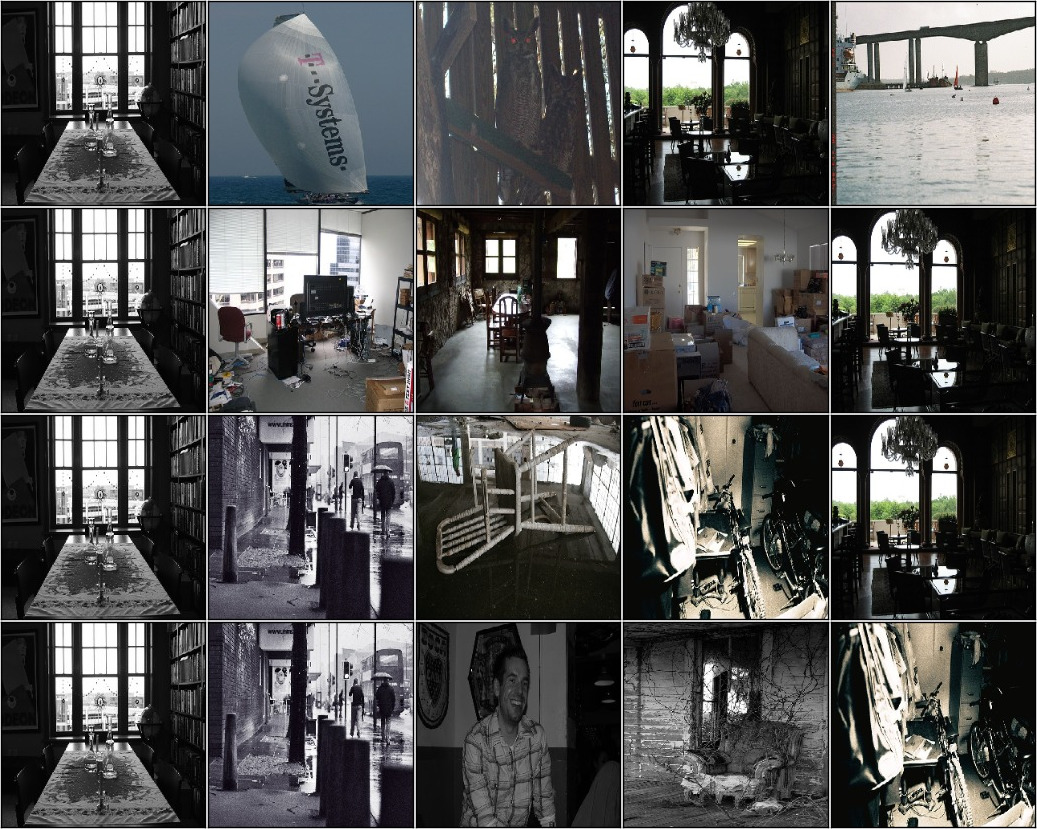}
  \hspace{1pt}
  \includegraphics[width=0.3\textwidth]{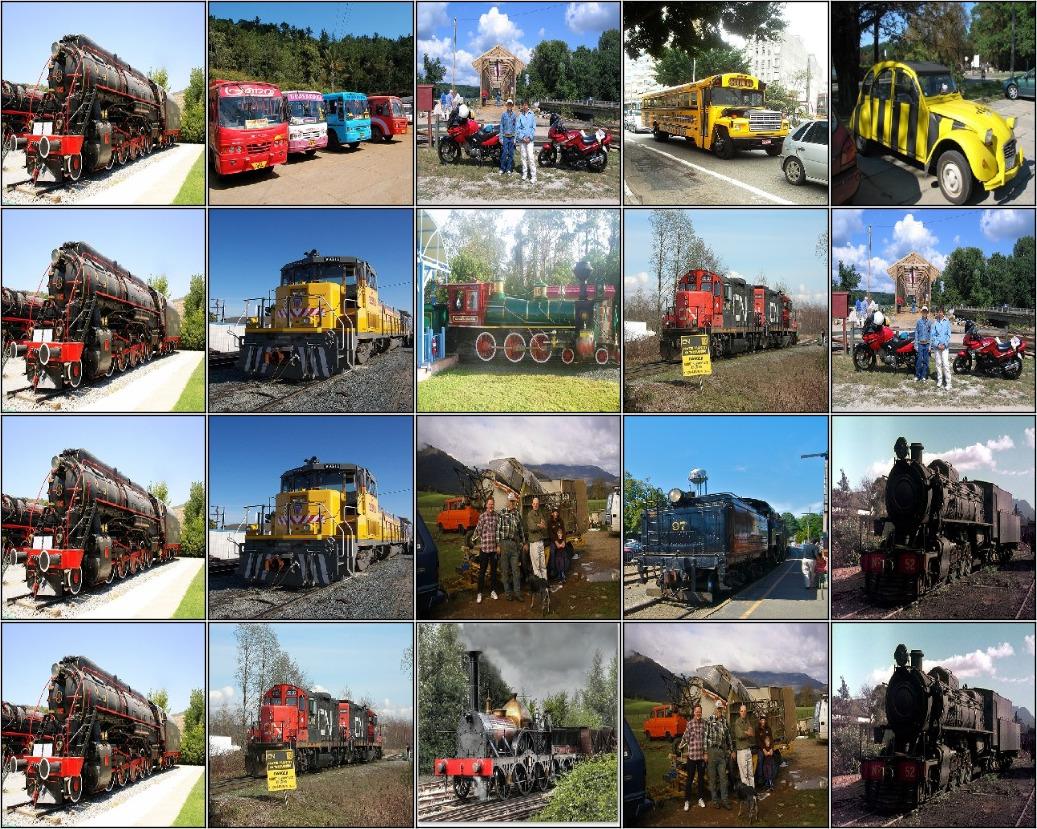}\\
  \vspace{4pt} 
  \includegraphics[width=0.3\textwidth]{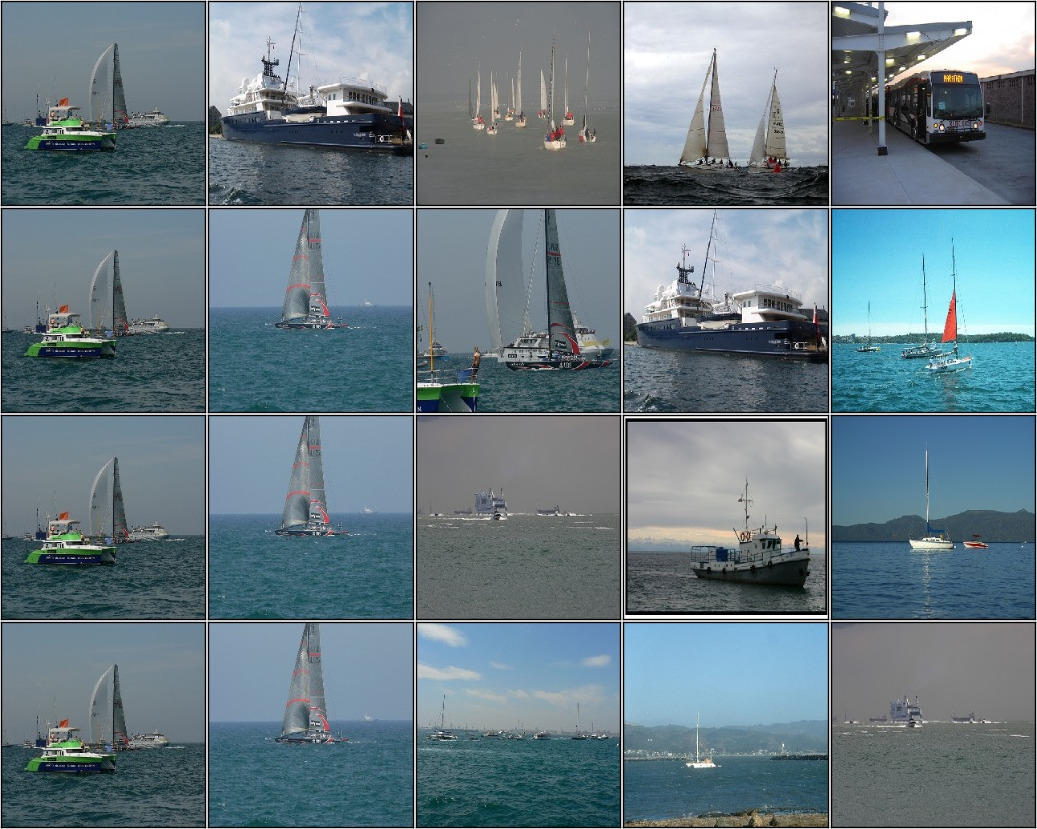}
  \hspace{1pt}
  \includegraphics[width=0.3\textwidth]{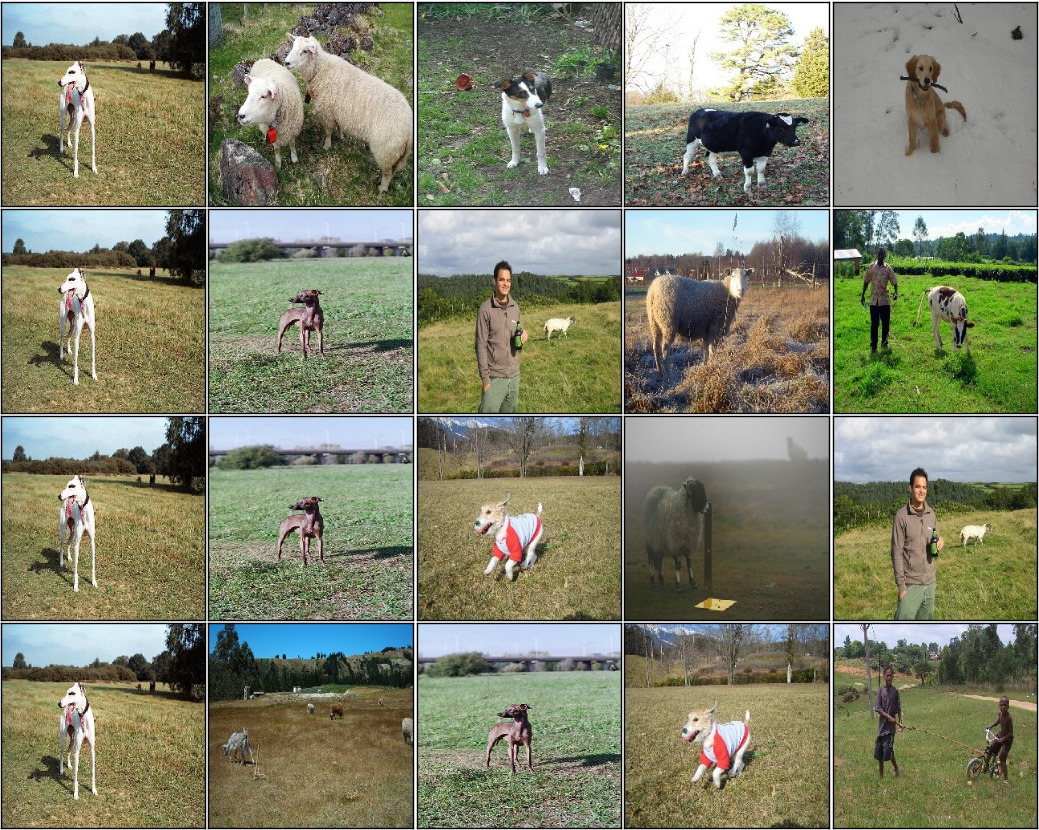}
  \hspace{1pt}
  \includegraphics[width=0.3\textwidth]{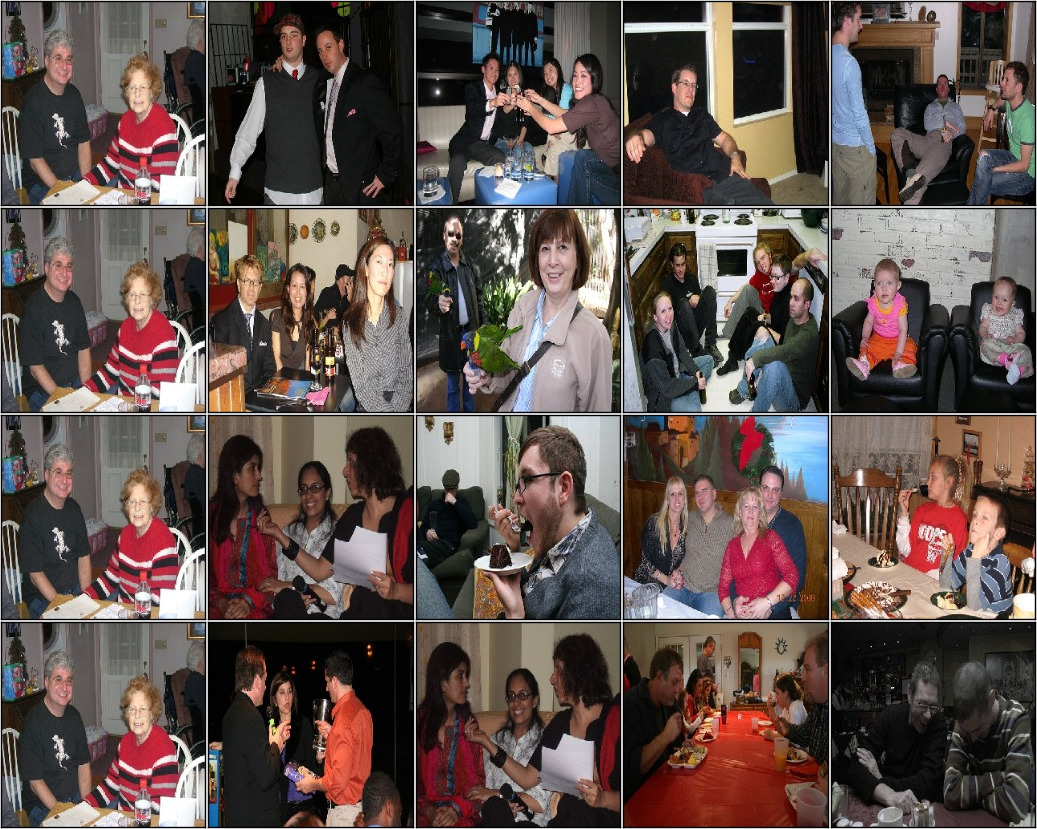}
%  \vspace{0.031em}\\
 %   \includegraphics[width=0.31\textwidth]{amb1_png.jpg}
 % \hfill
 % \includegraphics[width=0.31\textwidth]{amb3_png.jpg}
 % \hfill
 % \includegraphics[width=0.31\textwidth]{amb5_png.jpg}\\
  \caption{Top 4 nearest neighbors for a given query image image (left-most). Each row makes use of features obtained from different layers of TextTopicNet (without fine tuning). From top to bottom: prob, fc7, fc6, pool5.}
  \label{img2img}
\end{figure*}
\begin{figure*}
\centering
  \includegraphics[width=0.3\textwidth]{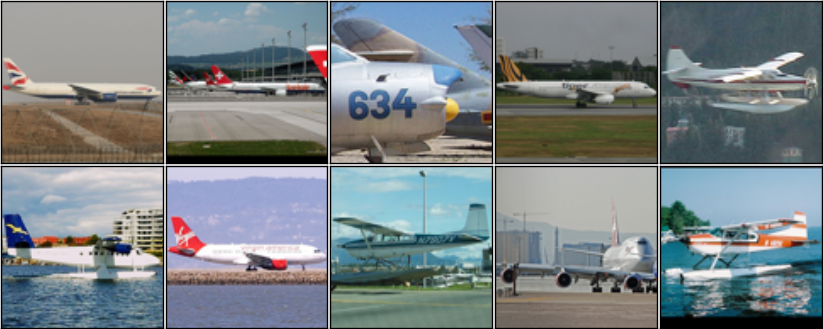}
  \hspace{1pt}
  \includegraphics[width=0.3\textwidth]{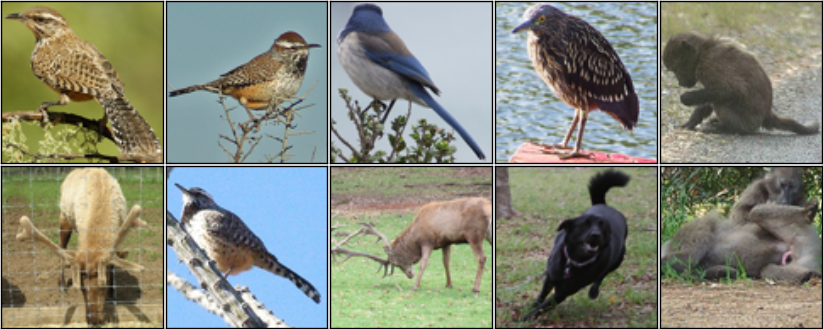}
  \hspace{1pt}
  \includegraphics[width=0.3\textwidth]{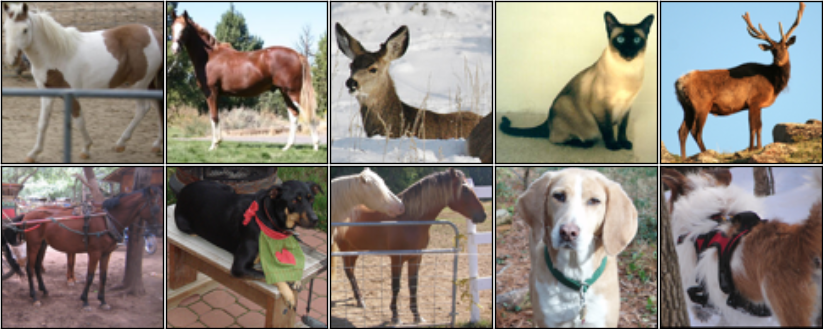}
  \caption{Top 10 nearest neighbors for a given text query (from left to right: ``airplane'', ``bird'', and ``horse'') in the topic space of TextTopicNet.}
  \label{txt2img}
\end{figure*}

\subsection{Multi-modal image retrieval}

{We evaluate our learned self-supervised visual features for two types of multi-modal retrieval tasks: (1) Image query vs. Text database, (2) Text query vs. Image database. For this purpose, we use the Wikipedia dataset \cite{rasiwasia2010new}, which consists of $2,866$ image-document pairs split into train and test set of $2173$ and $693$ pairs respectively. 
%Further, each image-document pair is labeled with one of ten semantic classes \cite{rasiwasia2010new}. 
%Text features are derived from the Latent Dirichlet Allocation model with 40 topics as explained in Section \ref{sec:train_lda}. Images are projected into text-topic space using the deep CNN we trained in self-supervised setting as explained in Section \ref{sec:train_cnn}. For retrieval, we 
For retrieval we project images and documents into the learned topic space and compute the KL-divergence distance of the query (image or text) with all the entities in the database. In Table \ref{table:multi_modal_retrieval} we compare our results with supervised and unsupervised multi-modal retrieval methods discussed in ~\cite{wang2016comprehensive} and ~\cite{kang2015cross}. Supervised methods make use of class or categorical information associated with each image-document pair, whereas unsupervised methods do not. All of these methods use LDA for text representation and CNN features from pre-trained CaffeNet \cite{jia2014caffe}, which is trained on ImageNet dataset \cite{deng2009imagenet} in a supervised setting. We appreciate that our self-supervised method outperforms unsupervised approaches, and has competitive performance to supervised methods without using any labeled data.
}

\begin{table}[h]
\centering
\begin{tabular}{l c c c}
\toprule
Method & Image query & Text query & Average \\
\midrule
TextTopicNet & $39.58$ & $38.16$ & $38.87$ \\
\midrule
CCA \cite{hardoon2004canonical,rasiwasia2010new}& $19.70$ & $17.84$ & $18.77$ \\
PLS \cite{rosipal2006overview} & $30.55$ & $28.03$ & $29.29$ \\
\midrule
SCM \cite{rasiwasia2010new}& $37.13$ & $28.23$ & $32.68$ \\
GMMFA \cite{sharma2012generalized} & $38.74$ & $31.09$ & $34.91$ \\
CCA-3V \cite{gong2014multi} & $40.49$ & $36.51$ & $38.50$ \\
GMLDA \cite{sharma2012generalized} & $40.84$ & $36.93$ & $38.88$ \\
LCFS \cite{wang2013learning}& $41.32$ & $38.45$ & $39.88$ \\
JFSSL \cite{wang2016joint}& $42.79$ & $39.57$ & $41.18$ \\
\bottomrule
\end{tabular}
\caption{MAP comparison on Wikipedia dataset \cite{rasiwasia2010new} with supervised (bottom) and unsupervised (middle) methods.}
\label{table:multi_modal_retrieval}
\end{table}
%-----%

Finally, in order to analyze better what is the nature of learned features by our self-supervised TextTopicNet we perform additional qualitative experiments for an image retrieval task. 

%Figure~\ref{img2img} shows the 25 nearest neighbors for a given query image image (blue frame) in the topic space layer (40 dimensions) of TextTopicNet (without fine tuning).

Figure~\ref{img2img} shows the 4 nearest neighbors for a given query image (left-most), where each row makes use of features obtained from different layers of TextTopicNet (without fine tuning). From top to bottom: prob, fc7, fc6, pool5. Query images are randomly selected from PASCAL VOC 2007 dataset and never shown at training time. It can be appreciated that when retrieval is performed in the topic space layer (prob, 40 dimensions, top row), the results are semantically close, although not necessarily visually similar. As features from earlier layers are used, the results tend to be more visually similar to the query image.

%Notice that since the LDA model allows us to project text queries into the topic space we can also do text based retrieval.
Figure~\ref{txt2img} shows the 10 nearest neighbors for a given text query (from left to right: ``airplane'', ``bird'', and ``horse'') in the topic space of TextTopicNet (again, without fine tuning). Interestingly, the list of retrieved images for the first query (``airplane'') is almost the same for related words and synonyms such as  ``flight'', ``airway'', or ``aircraft''. By leveraging textual semantic information our method learns a polysemic representation of images. 
% I like the idea of generation of word clouds from images as discussed with Dimos: From the input image you get the topic prob distribution and then you calculate a ranking of the most probable words for this image, then a tag cloud is created by taking N top words weighted by their ranking score. a visualization of the "conceptual" meaning of the image.

%\alert{LL: TODO: add here Figure with some "word clouds" generated by our method for a given image input. to extend a bit the idea about "polysemic" representations }

%\subsection{Scene Text lexicon generation}

%------------------------------------------------------------------------
\section{Conclusion}

In this paper we have presented a method that is able to take advantage of freely available multi-modal content to train computer vision algorithms without human supervision. By considering text found in illustrated articles as noisy image annotations the proposed method learns visual features by training a CNN to predict the semantic context in which a particular image is more probable to appear as an illustration. 

The contributed experiments show that although the learned visual features are generic for broad topics, they can be used for more specific computer vision tasks such as image classification, object detection, and multi-modal retrieval. Our results are comparable with state of the art self-supervised algorithms for visual feature learning.

{TextTopicNet source code and pre-trained models are publicly available at} \url{https://git.io/vSotz}.

\section*{Acknowledgment}
We gratefully acknowledge the support of the NVIDIA Corporation with the donation of the Titan X Pascal GPU used for this research. This work has been partially supported by the Spanish research project TIN2014-52072-P and the CERCA Programme/Generalitat de Catalunya. %This project was supported by the Spanish project TIN2011-24631. This project was supported by the Spanish project TIN2011-24631.

{\small
\bibliographystyle{ieee}
\bibliography{bibliography}
}

\end{document}